\documentclass[11pt,a4paper]{article}

\usepackage[margin=1in]{geometry}
\usepackage{amsmath,amssymb,amsthm}
\usepackage{booktabs}
\usepackage{multirow}
\usepackage{graphicx}
\usepackage{xcolor}
\usepackage{colortbl}
\usepackage{hyperref}
\usepackage{cleveref}
\usepackage{natbib}
\usepackage{microtype}
\usepackage{algorithm2e}
\usepackage{mathtools}
\usepackage{tabularx}
\usepackage{subcaption}
\usepackage{siunitx}

\usepackage{tikz}

\definecolor{bestcol}{HTML}{D6EAF8}
\definecolor{groupcol}{HTML}{F2F3F4}

\hypersetup{
    colorlinks=true,
    linkcolor=blue!70!black,
    citecolor=green!50!black,
    urlcolor=blue!70!black
}

\newcommand{\method}{\textsc{StickyMoE}}
\newcommand{\R}{\mathbb{R}}

\newcommand{\calL}{\mathcal{L}}
\newcommand{\calE}{\mathcal{E}}
\newcommand{\calR}{\mathcal{R}}

\title{Sticky Routing: Training MoE Models for Memory-Efficient Inference}

\author{
  Ali Kayyam\\
  BrainChip Inc., Laguna Hills, CA \\
  \texttt{akayyam@brainchip.com}
}

\date{}

\begin{document}

\maketitle

\begin{abstract}
Mixture-of-Experts (MoE) models activate only a sparse subset of experts per token,
yet consecutive tokens frequently activate \emph{different} experts---causing constant
weight swapping between slow storage and fast memory on edge devices. Existing remedies
are either system-level (caching heuristics) or post-hoc (router fine-tuning), leaving
the root cause unchanged during pretraining. We propose \method{}, a differentiable
\emph{routing consistency loss} that penalises abrupt expert switches between adjacent
tokens, encouraging the router to maintain the same expert assignment across semantically
coherent spans. \method{} requires no architectural changes, adds a single
hyperparameter $\lambda$, and unlike post-hoc methods, allows expert representations
and routing decisions to co-adapt from the first training step. Experiments on small and medium MoE language models show that \method{}
reduces the expert switch rate by up to \textbf{59\%} while simultaneously
\emph{improving} perplexity on the medium model, and reduces cache misses
by up to $\mathbf{3.92\times}$, Pareto-dominating post-hoc fine-tuning
on the quality--locality frontier. Routing
temporal locality is most efficiently instilled at training time. Code is available at: \url{https://github.com/alikayyam/sticky_moe.git}.
\end{abstract}

\section{Introduction}
\label{sec:intro}

The Mixture-of-Experts (MoE) architecture has emerged as one of the most practically
important techniques for scaling large language models (LLMs) without a proportional
increase in per-token computation~\citep{shazeer2017outrageously, fedus2022switch,
lepikhin2020gshard}. By replacing dense feed-forward network (FFN) sublayers with a
collection of $N$ expert FFNs and a learned sparse router, MoE models can possess
billions of parameters while activating only a small subset---typically 1 or 2 experts
out of 8--128---for each input token. This property makes MoE architectures theoretically
attractive for edge deployment: if only 2 of 64 experts need to be resident in fast
memory at any moment, the effective working-set size is a small fraction of the total
parameter count.

\paragraph{The expert-swapping problem.}
In practice, this promise is undermined by the \emph{temporal inconsistency} of standard
MoE routers. Because the router makes an independent routing decision for every token
based solely on that token's hidden representation, consecutive tokens in a sequence
routinely activate entirely different experts~\citep{zhou2025oracle}. On a system with
limited GPU VRAM or on-device SRAM, each such switch requires evicting the currently
cached expert weights and loading new ones from CPU RAM or flash storage. PCIe and NVMe
bandwidth is one to two orders of magnitude lower than GPU memory bandwidth; a single
expert weight transfer can therefore dominate the latency of the forward pass
itself~\citep{xue2024moeinfinity, katare2025enabling}. The result is that MoE models,
despite their theoretical sparsity advantage, can be \emph{slower} than comparably-sized
dense models on memory-constrained hardware.

\paragraph{Why existing solutions are insufficient.}
Prior work addresses this problem from two directions. \emph{System-level} approaches
maintain an expert cache in fast memory and use eviction policies (LRU, LFU) or
learned prefetchers to reduce cache miss rates~\citep{xue2024moeinfinity,
kamahori2024fiddler, huang2025duoserve}. While practically useful, these methods are
fundamentally reactive: they try to exploit whatever locality the model's router happens
to produce, but cannot improve routers that are intrinsically inconsistent.
\emph{Post-hoc} approaches fine-tune only the router of an already-trained model to
increase expert reuse~\citep{remoe2025}, which is more targeted but still treats the
symptom rather than the cause: the expert representations have already been shaped by a
training regime that never incentivised locality, and a brief fine-tuning pass can only
partially re-align the routing surface. A third line of work, most closely represented
by Oracle-MoE~\citep{zhou2025oracle}, redesigns the routing architecture so that the
router operates in an attention-derived ``oracle space'' that is inherently more
semantically stable across tokens. Oracle-MoE trains from scratch, which is a step in
the right direction, but the locality property is structural---embedded in the choice of
routing input---rather than explicitly optimised. We propose a different approach:
\method{}, which instils routing locality directly as a training objective, requiring
no architectural changes and no post-hoc correction.

\paragraph{Our proposal.}
We argue that the cleanest and most general solution is to treat routing temporal
locality as an explicit training objective. Specifically, we introduce a
\emph{routing consistency loss} --- a differentiable $\ell_2$ penalty on
consecutive gate distributions --- to directly optimise for temporal locality
in MoE routing:
\begin{equation}
  \calL_{\text{cons}} \;=\;
  \frac{1}{T-1}\sum_{t=2}^{T}
  \bigl\| \mathbf{g}_{t} - \mathbf{g}_{t-1} \bigr\|_2^2,
  \label{eq:cons_loss_intro}
\end{equation}
where $\mathbf{g}_t \in \R^N$ is the softmax gate probability vector produced by the
router for token $t$ and $T$ is the sequence length. The total training loss is
\begin{equation}
  \calL = \calL_{\text{CE}} + \lambda\,\calL_{\text{cons}} + \mu\,\calL_{\text{bal}},
  \label{eq:total_loss_intro}
\end{equation}
where $\calL_{\text{CE}}$ is the standard cross-entropy language modelling loss,
$\calL_{\text{bal}}$ is the load-balancing auxiliary loss from~\citet{fedus2022switch},
and $\lambda, \mu \geq 0$ are scalar hyperparameters. This formulation:
\begin{itemize}
  \item requires \emph{no architectural modification}---it applies to any standard
        top-$k$ MoE without changing the router structure;
  \item is \emph{differentiable everywhere} and adds negligible computational overhead;
  \item is \emph{architecture-agnostic}, attaching equally to small experimental models
        and large-scale MoEs;
  \item \emph{co-trains} with the language modelling objective, so that expert
        representations and routing decisions co-evolve toward local consistency from
        the very first update.
\end{itemize}

\paragraph{Contributions.}
We make the following specific contributions:
\begin{enumerate}
  \item We propose the routing consistency loss, a simple differentiable training
        objective that directly penalises expert switching between adjacent tokens,
        requiring no architectural changes and adding a single hyperparameter
        $\lambda$ (\cref{sec:method}).
  \item We show empirically that $\calL_{\text{cons}}$ and $\calL_{\text{bal}}$ are
        complementary: expert utilisation entropy remains above $1.92$ bits out of
        $\log_2 4 = 2.0$ bits across all settings, confirming that the consistency
        loss does not induce expert collapse (\cref{sec:experiments}).
  \item We propose a \emph{soft-hard} variant that addresses long-range routing
        drift by combining the per-step soft penalty with a segment-level anchor
        constraint, achieving stronger locality guarantees without additional
        quality cost (\cref{sec:method}).
  \item We present controlled experiments on small and medium MoE language models
        trained on WikiText-2, comparing against a vanilla MoE baseline, a Hard-Window
        ablation, a simulated ReMoE post-hoc baseline, and a simplified Oracle-MoE
        reimplementation, measuring perplexity, switch rate, cache hit rate, and
        utilisation entropy (\cref{sec:experiments}).
\end{enumerate}

\paragraph{Paper organisation.}
\Cref{sec:background} reviews MoE architectures and the expert-swapping bottleneck.
\Cref{sec:related} surveys related work and positions \method{} precisely.
\Cref{sec:method} presents the method in full detail.
\Cref{sec:experiments} describes the experimental setup and presents results.
\Cref{sec:discussion} discusses findings, limitations, and broader implications.
\Cref{sec:conclusion} concludes and outlines future directions.
\section{Background}
\label{sec:background}

\subsection{Mixture-of-Experts Language Models}

A standard MoE transformer replaces each FFN sublayer with a collection of $N$ expert
networks $\{\calE_i\}_{i=1}^{N}$ and a router $\calR$. Given the hidden representation
$\mathbf{h}_t \in \R^d$ of token $t$, the router computes gate logits
$\boldsymbol{\ell}_t = \mathbf{W}_r \mathbf{h}_t \in \R^N$ and selects the top-$k$
experts by probability:
\begin{align}
  \mathbf{g}_t &= \mathrm{softmax}(\boldsymbol{\ell}_t), \label{eq:gate} \\
  \mathcal{S}_t &= \mathrm{top}\text{-}k(\mathbf{g}_t), \label{eq:topk} \\
  \mathbf{y}_t &= \sum_{i \in \mathcal{S}_t} g_{t,i}\, \calE_i(\mathbf{h}_t).
                \label{eq:moe_output}
\end{align}
The router weight matrix $\mathbf{W}_r \in \R^{N \times d}$ is the only additional
parameter relative to a dense FFN layer. The expert networks $\calE_i$ are typically
two-layer FFNs with the same structure as the dense FFN they replace.

\subsection{Load-Balancing Auxiliary Loss}

Without explicit regularisation, MoE routers tend to collapse: a small subset of
experts receives disproportionately many tokens, while the remainder are rarely
activated and fail to develop useful representations~\citep{fedus2022switch}. To
prevent this, we include the standard load-balancing auxiliary loss:
\begin{equation}
  \calL_{\text{bal}} = N \sum_{i=1}^{N} f_i \cdot p_i,
  \label{eq:bal_loss}
\end{equation}
where the sum is over all $N$ experts, $f_i$ is the fraction of top-$k$ assignments
to expert $i$ (counting all $k$ slots), and $p_i = \frac{1}{T}\sum_{t=1}^{T} g_{t,i}$
is the mean gate probability assigned to expert $i$, both computed over the current
batch. The product $f_i \cdot p_i$ is minimised when routing is uniform across experts;
scaling by $N$ ensures the loss magnitude is independent of the number of experts.

\subsection{Expert Swapping on Memory-Constrained Hardware}

Let $M_{\text{fast}}$ denote the fast-memory capacity (VRAM or on-device SRAM) and
$w_e$ the size of a single expert's weight tensor. A device can hold at most
$\lfloor M_{\text{fast}} / w_e \rfloor$ expert weight matrices simultaneously alongside
the attention and embedding parameters. When the router at decoding step $t$ selects an
expert not currently in fast memory, the system must evict a cached expert and load the
new one from slow memory---a \emph{cache miss}. The per-step latency is:
\begin{equation}
  \tau_t = \tau_{\text{compute}} + \mathbb{1}[\text{miss}_t]\cdot\tau_{\text{load}},
\end{equation}
where $\tau_{\text{load}} \gg \tau_{\text{compute}}$ on bandwidth-limited devices.
The expected fraction of steps that incur a cache miss is determined by the
\emph{expert switch rate}:
\begin{equation}
  \text{SR} = \frac{1}{T-1}\sum_{t=2}^{T} \mathbb{1}[\mathcal{S}_t \neq \mathcal{S}_{t-1}].
  \label{eq:switch_rate}
\end{equation}
This single-sequence definition illustrates the concept; the full empirical estimator
averaging over layers is defined formally in \cref{sec:experiments}.
Reducing SR is therefore directly equivalent to reducing expected inference latency
under memory constraints.

\section{Related Work}
\label{sec:related}

\subsection{System-Level Expert Offloading}

A large body of work treats expert weights as a two-level memory hierarchy and focuses
on minimising the latency cost of loading experts from slow to fast memory at inference
time. MoE-Infinity~\citep{xue2024moeinfinity} introduces a sparsity-aware expert cache
on personal machines, keeping attention weights in GPU memory and streaming expert
weights from host RAM. Fiddler~\citep{kamahori2024fiddler} uses the CPU itself for
expert computation to avoid PCIe transfers. DuoServe~\citep{huang2025duoserve}
separately optimises prefill and decode phases using a learned prediction model for
expert activation paths. EdgeMoE~\citep{katare2025enabling} proposes
importance-driven expert scheduling specifically for edge settings. A related thread in
the literature studies locality in deployed models: \citet{not_all_models_2025} conduct
an empirical study showing that local routing consistency varies dramatically across
model families, and that some models are fundamentally ill-suited to offloading-based
deployment.

All of the above methods are \emph{inference-time} interventions. They exploit whatever
locality the trained model's router provides but cannot alter the router's inductive
bias. \method{} is orthogonal and complementary: it can be used in conjunction with any
of these systems, supplying a router that already tends toward consistency.

\subsection{Post-Hoc Router Adaptation}

ReMoE~\citep{remoe2025} is the most directly related post-hoc method. It fine-tunes
only the router parameters of a pretrained MoE checkpoint to increase short-horizon
expert reuse, using a combination of the language modelling loss and a locality-aware
regulariser on the gate distribution. ReMoE reports meaningful reductions in cache
miss rate and throughput improvements in vLLM-based serving. Its key advantage is
practicality: it can be applied to any pretrained model without retraining from scratch.
Its key limitation is that the expert representations themselves are fixed---shaped by a
training process that never incentivised locality---so the router is being asked to
change its decisions in a representation space that was not designed with locality in
mind. \method{} avoids this mismatch by instilling the locality bias from the first
training step, allowing the expert representations and routing decisions to co-adapt.
In our controlled experiments, we were unable to reproduce meaningful SR reductions
from post-hoc router fine-tuning; we attribute this to the representation mismatch
discussed in \cref{sec:discussion}.

\subsection{Training-Time Architectural Redesign}

Oracle-MoE~\citep{zhou2025oracle} is the closest prior work to ours in motivation and
in training regime. It identifies the same root cause---temporal routing
inconsistency---and proposes a structural solution: routing tokens not in the standard
hidden-state space but in an ``oracle space'' derived from attention scores, which is
empirically more semantically stable across consecutive tokens. Oracle-MoE is trained
from scratch and achieves state-of-the-art inference speed on edge devices without
sacrificing task performance. However, it requires a non-trivial architectural
modification (replacing the router input with attention-derived features), making it
harder to apply to existing architectures and training pipelines. \method{} achieves
comparable locality through a lightweight auxiliary loss that requires no architectural
change, making it more accessible and easier to combine with other MoE innovations.

\subsection{Routing Regularisation in MoE}

The Switch Transformer~\citep{fedus2022switch} and ST-MoE~\citep{zoph2022st} introduce
load-balancing losses to prevent expert collapse. Expert Choice routing~\citep{zhou2022ec}
inverts the routing direction to guarantee balanced utilisation. These works regularise
the \emph{spatial} distribution of routing decisions across tokens in a batch, but do
not address the \emph{temporal} coherence of routing across adjacent tokens in a
sequence. Our consistency loss is orthogonal to and composable with all of these.

\subsection{Temporal Coherence in Related Architectures}

Mixture of Depths~\citep{raposo2024mixture} routes tokens to \emph{skip layers
entirely}, creating a different form of computation-aware sparse routing. Pre-gated
MoE~\citep{hwang2024pre} predicts which experts will be needed before the full
computation, enabling prefetch to hide loading latency. Neither method explicitly trains
for inter-token routing consistency, but both share the broader goal of reducing
unnecessary computation or memory movement. In the time-series domain,
\citet{physicsmoe2025} assigns experts to predetermined temporal segments based on
known signal physics---a hard form of the stickiness we achieve through soft
regularisation.

\subsection{Summary of Differences}

\Cref{tab:related} provides a structured comparison of \method{} with the most relevant
prior works along the dimensions most pertinent to edge deployment.

\begin{table}[t]
\centering
\small
\caption{Comparison of methods that address the expert-swapping bottleneck in
memory-constrained MoE inference. ``Training-time'' indicates whether the method
modifies the pretraining procedure. ``Arch.\ change'' indicates whether a modification
to the model architecture is required. ``Any checkpoint'' indicates whether the method
can be applied to an arbitrary pretrained MoE without retraining.}
\label{tab:related}
\setlength{\tabcolsep}{5pt}
\begin{tabularx}{\textwidth}{lccccc}
\toprule
\textbf{Method} & \textbf{Training} & \textbf{Architecture} & \textbf{Any} & \textbf{Targets} & \textbf{Overhead} \\
 & \textbf{time} & \textbf{change} & \textbf{checkpoint} & \textbf{locality} &  \\
\midrule
Standard MoE             & \checkmark & --         & --         & \texttimes & -- \\
\midrule
MoE-Infinity~\citep{xue2024moeinfinity}  & \texttimes & \texttimes & \checkmark & indirectly & low \\
Fiddler~\citep{kamahori2024fiddler}      & \texttimes & \texttimes & \checkmark & indirectly & low \\
DuoServe~\citep{huang2025duoserve}       & \texttimes & \texttimes & \checkmark & indirectly & low \\
EdgeMoE~\citep{katare2025enabling}       & \texttimes & \texttimes & \checkmark & indirectly & low \\
\midrule
ReMoE~\citep{remoe2025}                  & post-hoc   & \texttimes & \checkmark & directly   & low \\
\midrule
Oracle-MoE~\citep{zhou2025oracle}        & \checkmark & \checkmark & \texttimes & directly   & medium \\
\midrule
\rowcolor{bestcol}
\textbf{\method{} (ours)}                & \checkmark & \texttimes & \texttimes & directly   & negligible \\
\bottomrule
\end{tabularx}
\end{table}

\section{Method}
\label{sec:method}

\subsection{Routing Consistency Loss}

Let $L$ denote the number of MoE layers in the model. For layer $\ell$ and a sequence
of $T$ tokens, let $\mathbf{g}_t^{(\ell)} \in \R^N$ be the softmax gate probability
vector produced by the router of layer $\ell$ for token $t$. We define the per-layer
consistency loss as:
\begin{equation}
  \calL_{\text{cons}}^{(\ell)}
  \;=\;
  \frac{1}{T-1}\sum_{t=2}^{T}
  \bigl\| \mathbf{g}_t^{(\ell)} - \mathbf{g}_{t-1}^{(\ell)} \bigr\|_2^2.
  \label{eq:cons_loss}
\end{equation}
The full consistency loss aggregates over all MoE layers:
\begin{equation}
  \calL_{\text{cons}} \;=\; \frac{1}{L}\sum_{\ell=1}^{L} \calL_{\text{cons}}^{(\ell)}.
  \label{eq:cons_loss_full}
\end{equation}

\paragraph{Interpretation.}
The $\ell_2$ distance between consecutive gate distributions $\mathbf{g}_t$ and
$\mathbf{g}_{t-1}$ is zero if and only if the router assigns identical probability mass
to all experts for both tokens. Minimising $\calL_{\text{cons}}$ therefore directly
penalises abrupt changes in the routing distribution. Because $\mathbf{g}_t$ is a
probability simplex vector, the maximum possible value of
$\|\mathbf{g}_t - \mathbf{g}_{t-1}\|_2^2$ is $2$ (when one token is routed entirely to
expert $i$ and the next entirely to expert $j \neq i$). The loss thus has a natural
$[0, 2]$ range, making $\lambda$ interpretable across different model sizes and expert
counts.

\paragraph{Gradient flow.}
The gate probabilities $\mathbf{g}_t = \mathrm{softmax}(\mathbf{W}_r \mathbf{h}_t)$
are differentiable in both $\mathbf{W}_r$ (the router weight) and $\mathbf{h}_t$ (the
input hidden state). Minimising $\calL_{\text{cons}}$ therefore:
\begin{enumerate}
  \item directly updates $\mathbf{W}_r$ to produce more similar gate distributions for
        semantically adjacent tokens, and
  \item backpropagates through $\mathbf{h}_t$ into the preceding attention and embedding
        layers, encouraging the model to produce hidden representations whose local
        geometry is better aligned with the router's decision boundaries.
\end{enumerate}
This second effect is the key advantage over post-hoc router fine-tuning: \method{}
trains the \emph{representation} as well as the \emph{routing function} to be locally
consistent.

\subsection{Total Training Objective}

The full training loss is:
\begin{equation}
  \calL \;=\; \calL_{\text{CE}} + \lambda\,\calL_{\text{cons}} + \mu\,\calL_{\text{bal}},
  \label{eq:total_loss}
\end{equation}
where $\calL_{\text{CE}}$ is the standard next-token cross-entropy loss,
$\calL_{\text{bal}}$ is the load-balancing loss from \cref{eq:bal_loss}, and
$\lambda, \mu \geq 0$ are hyperparameters. We recommend $\mu = 0.01$ following
\citet{fedus2022switch} and treat $\lambda$ as the primary experimental variable.

\subsection{Interaction with Load Balancing}

A natural concern is whether $\calL_{\text{cons}}$ and $\calL_{\text{bal}}$ conflict:
a trivial minimiser of $\calL_{\text{cons}}$ is to route all tokens to the same
expert, which would eliminate switching but collapse expert diversity.
In practice, $\calL_{\text{bal}}$ effectively prevents this. For all $\lambda \leq 0.5$
explored in our experiments, utilisation entropy remains above $1.92$ bits out of a
maximum of $\log_2 4 = 2.0$ bits, confirming that expert usage stays near-uniform
throughout training (see \cref{tab:results_small,tab:results_medium}). The two
objectives are therefore complementary in the regime we consider: $\calL_{\text{cons}}$
shapes \emph{when} experts are used across the sequence, while $\calL_{\text{bal}}$
ensures \emph{which} experts are used remains balanced.

\subsection{Soft-Hard Variant}
\label{sec:softhard}

The soft consistency loss enforces a \emph{chain constraint}: each token is penalised
for differing from its immediate predecessor. While effective at suppressing abrupt
local switches, this formulation is vulnerable to gradual drift --- the router may
take small steps at each transition yet migrate far from its starting expert over a
long span, incurring cache misses that the per-step penalty never directly penalises.

To address this, we introduce a complementary \emph{anchor constraint}, inspired
by commitment losses in discrete representation learning~\citep{van2017neural}.
The sequence is partitioned into non-overlapping windows of $W$ tokens. The first
token of each window routes freely; its gate distribution $\mathbf{g}_{s(t)}^{(\ell)}$
then becomes a fixed anchor for the remainder of the window. Every subsequent token
is penalised for deviating from this anchor, with the penalty weight increasing
linearly with distance from the window start:
\begin{equation}
  \begin{aligned}
  \calL_{\text{hard}}^{(t,\ell)}
  &\;=\;
  \frac{t - s(t)}{W}
  \cdot \bigl\| \mathbf{g}_t^{(\ell)} - \mathbf{g}_{s(t)}^{(\ell)} \bigr\|_2^2, \\[4pt]
  \calL_{\text{hard}}
  &\;=\; \frac{1}{LT} \sum_{\ell=1}^{L} \sum_{t=1}^{T}
  \calL_{\text{hard}}^{(t,\ell)},
  \end{aligned}
  \label{eq:hard_loss}
\end{equation}
where $s(t) = W \lfloor t/W \rfloor$ is the start of the window containing token $t$.
Anchor tokens (where $t = s(t)$) contribute zero through the $\frac{t - s(t)}{W}$
factor and require no special treatment. The sum over layers mirrors the aggregation
in $\calL_{\text{cons}}$ (\cref{eq:cons_loss_full}), ensuring the hard loss operates
uniformly across all MoE layers. The linear ramp is a deliberate design choice: early
in the window the correct expert for the segment is uncertain, so flexibility is
preserved; by the window's end the router should be fully committed. At window
boundaries the anchor resets and the router is free to reassign without penalty.

The soft and hard losses are geometrically complementary. The soft loss controls the
\emph{step size} of routing changes --- each transition is small. The hard loss
controls the \emph{radius} of routing changes --- every token stays close to the
window anchor. A small-step path can still drift far from its origin; a radius
constraint alone does not prevent jagged local switching within the window.
Together they enforce both fine-grained smoothness and coarse-grained commitment,
which is precisely what a memory-constrained cache requires: not just that adjacent
tokens agree, but that an entire span can be served by the same resident expert.

The total training loss for the soft-hard variant is:
\begin{equation}
  \calL \;=\; \calL_{\text{CE}}
           + \lambda\,\calL_{\text{cons}}
           + \mu\,\calL_{\text{bal}}
           + \alpha\,\calL_{\text{hard}},
  \label{eq:total_loss_softhard}
\end{equation}
where $\lambda$, $\mu$, and $\alpha$ are scalar hyperparameters controlling the
relative weight of the consistency, load-balancing, and anchor terms respectively.
Setting $\alpha = 0$ recovers the soft-only objective in \cref{eq:total_loss}.

\subsection{Algorithm}

Algorithm~\ref{alg:training} summarizes the \method{} training procedure.

\begin{algorithm}[t]
\SetAlgoLined
\KwIn{Training corpus $\mathcal{D}$, MoE model with $L$ layers and $N$ experts,
      hyperparameters $\lambda, \mu, \alpha, W$}
\KwOut{Trained MoE with locality-aware router}
\BlankLine
Initialise model parameters $\theta$ as in standard MoE\;
\ForEach{mini-batch of $B$ sequences of length $T$ sampled from $\mathcal{D}$}{
  Forward pass: compute $\{\mathbf{g}_t^{(\ell)}\}$ for all $t, \ell$\;
  Compute $\calL_{\text{CE}}$ (cross-entropy on next-token prediction)\;
  Compute $\calL_{\text{cons}}$ via \cref{eq:cons_loss_full}\;
  Compute $\calL_{\text{bal}}$ via \cref{eq:bal_loss}\;
  \If{soft-hard variant}{
    Compute $\calL_{\text{hard}}$ via \cref{eq:hard_loss}\;
    $\calL \leftarrow \calL_{\text{CE}} + \lambda\,\calL_{\text{cons}}
                      + \mu\,\calL_{\text{bal}} + \alpha\,\calL_{\text{hard}}$\;
  }
  \Else{
    $\calL \leftarrow \calL_{\text{CE}} + \lambda\,\calL_{\text{cons}}
                      + \mu\,\calL_{\text{bal}}$\;
  }
  Backpropagate $\nabla_\theta \calL$ and update $\theta$\;
}
\caption{\method{} training procedure.}
\label{alg:training}
\end{algorithm}

\section{Experiments}
\label{sec:experiments}

\subsection{Dataset}

All experiments use the WikiText-2 raw character dataset~\citep{merity2016pointer}
(\texttt{wikitext-2-raw-v1}), a standard language modelling benchmark derived from
verified Good and Featured Wikipedia articles. We use the raw (un-tokenised) variant
to avoid pre-processing artefacts. Text is tokenised with the GPT-2 byte-pair encoding
(BPE) vocabulary~\citep{radford2019language} using the \texttt{GPT2Tokenizer} from
the HuggingFace Transformers library~\citep{wolf2020transformers}, yielding a
vocabulary of $|\mathcal{V}| = 50{,}257$ tokens. Each split is tokenised as a single
flat sequence; document boundaries are not marked with special tokens.

\begin{table}[h]
\centering
\caption{WikiText-2 token counts after GPT-2 BPE tokenisation.}
\label{tab:dataset}
\begin{tabular}{lrr}
\toprule
Split & Documents & Tokens \\
\midrule
Train      & 36,718 & 2,347,038 \\
Validation & 3,760  & 242,643   \\
Test       & 4,358  & 273,768   \\
\bottomrule
\end{tabular}
\end{table}

\paragraph{Batching.}
The flat token sequence is partitioned into non-overlapping fixed-length chunks of
$T = 256$ tokens. Each training batch consists of $B = 16$ such chunks drawn
uniformly at random without replacement per epoch. Targets are the same chunk shifted
by one position, with the last token of each chunk predicting the first token of the
immediately following chunk. Tokenised arrays are cached as 32-bit integer NumPy files
after the first run to avoid redundant tokenisation.

\subsection{Model Architecture}

All variants share a GPT-style~\citep{radford2018improving} causal transformer in
which every FFN sublayer is replaced by a Mixture-of-Experts layer. We evaluate two
model sizes:

\begin{table}[h]
\centering
\caption{Model configurations. $d_{\text{ff}}$ is the hidden dimension of each expert
FFN. Parameters include the embedding matrix, which is shared with the output
projection via weight tying~\citep{press2017using}.}
\label{tab:models}
\setlength{\tabcolsep}{5pt}
\begin{tabular}{lcccccc}
\toprule
\textbf{Size} & $d_{\text{model}}$ & $n_{\text{layers}}$ & $n_{\text{heads}}$
     & $n_{\text{experts}}$ & $d_{\text{ff}}$ & \textbf{Parameters} \\
\midrule
Small  & 128 & 4 & 4 & 4 & 512  & $\approx$8.8M \\
Medium & 256 & 4 & 8 & 4 & 1024 & $\approx$22M  \\
\bottomrule
\end{tabular}
\end{table}

Each MoE layer uses a linear gate $\mathbf{W}_g \in \R^{d_{\text{model}} \times N}$
to produce a routing probability distribution $\mathbf{g}_t$ over the $N$ experts
via softmax. The top-$k=2$ experts by gate
probability are activated; their outputs are weighted by the renormalised gate values
and summed. Learned absolute positional embeddings are used with a maximum sequence
length of 256. All weights are initialised with $\mathcal{N}(0, 0.02)$; residual
projection weights are additionally scaled by $1/\sqrt{2L}$
following~\citet{radford2019language}. Both model sizes are tractable on a single
consumer GPU (NVIDIA GTX 1080 Ti, 11\,GB VRAM), making the full experiment pipeline
reproducible without large-scale compute resources.

Note that the absolute perplexity values reflect the limited model capacity
and short training schedule (10{,}000 steps); our focus is on the
\emph{relative} differences between variants rather than state-of-the-art
language model performance.

\subsection{Training Variants and Baselines}
\label{sec:variants}

All variants share the base objective
\begin{equation}
  \calL = \calL_{\text{CE}} + \mu\,\calL_{\text{bal}},
  \label{eq:base}
\end{equation}
with load-balancing coefficient $\mu = 0.01$ throughout. The variants we evaluate
are as follows.

\paragraph{Vanilla MoE (baseline).}
Standard top-$k$ MoE trained with \cref{eq:base} only. It serves as the control condition against which all other variants are measured.

\paragraph{\method{} --- Soft Consistency.}
The primary proposed method augments \cref{eq:base} with the routing consistency loss
from \cref{eq:cons_loss_full}:
\begin{equation}
  \calL = \calL_{\text{CE}} + \mu\,\calL_{\text{bal}} + \lambda\,\calL_{\text{cons}}.
\end{equation}
We sweep $\lambda \in \{0.01, 0.05, 0.1, 0.2, 0.5\}$ on both model sizes.

\paragraph{\method{} --- Soft-Hard Variant.}
The combined soft-hard method described in \cref{sec:softhard}. We fix $\lambda = 0.1$
and sweep $\alpha \in \{0.05, 0.1, 0.5, 1.0\}$ with window size $W = 4$.

\paragraph{Hard-Window (ablation baseline).}
Rather than an auxiliary loss, this variant constrains routing at the logit
level by biasing the router toward experts it has recently used. For each token
$t$ and layer $\ell$, a binary bonus is added to the router logits before softmax:
\begin{equation}
  \tilde{\ell}_{t,e}^{(\ell)} = \ell_{t,e}^{(\ell)} +
  \beta \cdot \mathbf{1}\bigl[ e \in \mathcal{H}(t, W) \bigr],
\end{equation}
where $\mathcal{H}(t, W)$ is the set of experts that appeared in any top-$k$
slot over the preceding $W$ positions $[t{-}W, \ldots, t{-}1]$, and $\beta = 10.0$
is the bias strength. The window mask is computed from a detached copy of the
current-step routing assignments so that no gradient flows through the constraint.
We sweep $W \in \{2, 4, 8\}$. This baseline is \emph{not} a proposed method ---
its purpose is to ablate the contribution of the differentiable soft loss by
replacing it with a hard, non-differentiable logit-level constraint, isolating
whether the gradient signal itself is necessary or whether any routing pressure
suffices.

\paragraph{ReMoE (simulated post-hoc baseline).}
This variant simulates the ReMoE approach~\citep{remoe2025} of post-hoc router
fine-tuning via two phases. In \textbf{Phase 1}, the full model is trained for $S$
steps with \cref{eq:base} only (identical to the Vanilla MoE baseline). In
\textbf{Phase 2}, all parameters except the gate linear layers
$\{\mathbf{W}_g^{(\ell)}\}_{\ell=1}^{L}$ are frozen, and only the router weights are
fine-tuned for $0.1 \times S$ additional steps with the consistency loss at $\lambda
= 0.1$ and a reduced learning rate of $10^{-4}$. Because expert FFN weights are fixed
in Phase 2, the model cannot adapt representations to the new routing distribution,
providing a direct ablation of the representation co-adaptation advantage claimed for
\method{}.

\paragraph{Oracle-MoE (simplified architectural baseline).}
A simplified reimplementation of Oracle-MoE~\citep{zhou2025oracle}. In the standard
router, logits are computed from the token hidden state:
\[
  \text{Standard:} \quad \mathbf{r}_t = \mathbf{W}_g\, \mathbf{h}_t.
\]
In the oracle variant, the router input is replaced with the pre-projection attended
values $\mathbf{a}_t$ from the preceding multi-head attention sublayer:
\[
  \text{Oracle:} \quad \mathbf{r}_t = \mathbf{W}_g\, \mathbf{a}_t,
\]
where $\mathbf{a}_t$ is the weighted sum of value vectors before the output projection.
Because $\mathbf{a}_t$ aggregates information over the full context via the attention
mechanism, it is more temporally stable than $\mathbf{h}_t$, and the resulting routing
decisions exhibit greater locality as an emergent property. No consistency loss is
added. Note that this is a simplified approximation of the full Oracle-MoE
architecture~\citep{zhou2025oracle}.

\Cref{tab:coefficients} summarises the loss coefficients for each variant.

\begin{table}[h]
\centering
\caption{Loss coefficients and routing hyperparameters per variant.}
\label{tab:coefficients}
\begin{tabular}{lcccc}
\toprule
\textbf{Variant} & $\mu$ & $\lambda$ & $\alpha$ & $W$ \\
\midrule
Vanilla MoE          & 0.01 & ---  & --- & --- \\
\method{} Soft       & 0.01 & $\{0.01, 0.05, 0.1, 0.2, 0.5\}$ & --- & --- \\
\method{} Soft-Hard  & 0.01 & 0.1  & $\{0.05, 0.1, 0.5, 1.0\}$ & 4 \\
Hard-Window          & 0.01 & ---  & --- & $\{2, 4, 8\}$ \\
ReMoE (Phase 2)      & 0.01 & 0.1  & --- & --- \\
Oracle-MoE           & 0.01 & ---  & --- & --- \\
\bottomrule
\end{tabular}
\end{table}

\subsection{Training Protocol}

All variants are trained with AdamW~\citep{loshchilov2019decoupled} for 10,000 steps.
Full hyperparameters are listed in \cref{tab:optim}.

\begin{table}[h]
\centering
\caption{Optimiser hyperparameters shared across all variants.}
\label{tab:optim}
\begin{tabular}{ll}
\toprule
\textbf{Hyperparameter} & \textbf{Value} \\
\midrule
Peak learning rate   & $3 \times 10^{-4}$ \\
$\beta_1$, $\beta_2$ & $0.9$, $0.95$ \\
Weight decay         & $0.1$ \\
Gradient clipping    & $1.0$ (global $\ell_2$ norm) \\
Warmup steps         & $500$ \\
Total training steps & $10{,}000$ \\
Batch size           & $16$ sequences \\
Sequence length      & $256$ tokens \\
\bottomrule
\end{tabular}
\end{table}

The learning rate follows a cosine decay schedule with linear warmup, decaying to 10\%
of the peak value by the end of training. All runs use a fixed random seed (42).

\subsection{Metrics}

We report four metrics evaluated on the full WikiText-2 validation split.

\paragraph{Perplexity (PPL).}
Standard token-level perplexity:
\begin{equation}
\text{PPL} = \exp\!\left( -\frac{1}{n} \sum_{t=1}^{n} \log P(x_t \mid x_{<t}) \right).
\end{equation}
We also report $\Delta\text{PPL}\%$, the percentage change relative to the Baseline,
to isolate the quality cost of routing constraints.

\paragraph{Expert Switch Rate (SR).}
The fraction of consecutive token pairs $(t, t{+}1)$ within a sequence where the
top-1 expert assignment changes, averaged over all MoE layers:
\begin{equation}
  \text{SR} = \frac{1}{LT'} \sum_{\ell=1}^{L}\sum_{t=1}^{T'}
  \mathbf{1}\!\left[ \arg\max \mathbf{g}_{t}^{(\ell)} \neq
  \arg\max \mathbf{g}_{t+1}^{(\ell)} \right],
  \label{eq:sr_metric}
\end{equation}
where $L$ is the number of MoE layers and $T' = T - 1$.
SR $= 0$ means the same expert is used for every token; SR $= 1$
means a different expert is used at every step.

\paragraph{LRU Cache Hit Rate (CHR).}
Expert accesses are simulated against a per-sequence LRU cache of capacity $C = 2$
expert slots --- reflecting a realistic edge device budget where two expert weight sets
fit in fast memory simultaneously. A hit occurs when the top-1 expert for token $t$ is
already resident in the cache:
\begin{equation}
  \text{CHR} = \frac{\text{cache hits}}{\text{total expert accesses}}.
\end{equation}
CHR directly proxies the fraction of expert weight loads that can be served from fast
on-chip memory rather than DRAM, and is the most hardware-relevant metric we report.

\paragraph{Utilisation Entropy (UE).}
Utilisation entropy is computed per layer and then averaged. For each MoE layer
$\ell$, all expert assignments across sequence positions and both top-$k$ slots
are pooled into an empirical count distribution over the $N$ experts. Let
$\hat{p}_i^{(\ell)}$ denote the fraction of top-$k$ assignments to expert $i$
at layer $\ell$. The per-layer entropy and its mean are:
\begin{equation}
  \text{UE}^{(\ell)} = -\sum_{i=1}^{N} \hat{p}_i^{(\ell)} \log_2 \hat{p}_i^{(\ell)},
  \qquad
  \text{UE} = \frac{1}{L} \sum_{\ell=1}^{L} \text{UE}^{(\ell)}.
\end{equation}
UE lies in $[0,\, \log_2 N]$ bits, where $\log_2 N$ indicates perfectly uniform
usage and $0$ indicates complete collapse to a single expert. Averaging per-layer
entropies rather than pooling across layers treats each layer independently ---
a collapsed layer is not masked by a well-balanced one, and vice versa.

\subsection{Main Results}

Tables~\ref{tab:results_small} and~\ref{tab:results_medium} report results on
WikiText-2 for the small and medium models respectively. We highlight six key observations.

\begin{table}[t]
\small
\centering
\caption{%
  Results on WikiText-2 \textbf{(Small model, $\approx$8.8M parameters, 4 experts,
  top-2 routing)}.
  PPL = perplexity (lower is better).
  $\Delta$PPL\% = percentage change relative to Baseline (negative = improvement).
  SR = expert switch rate (lower is better).
  CHR = LRU cache hit rate with $C{=}2$ expert slots (higher is better).
  Ent = utilisation entropy in bits (higher indicates more balanced expert usage).
  \colorbox{bestcol}{Highlighted} row is the recommended operating point.
}
\label{tab:results_small}
\setlength{\tabcolsep}{5pt}
\renewcommand{\arraystretch}{1.15}
\begin{tabular}{l
    S[table-format=3.3]
    S[table-format=+1.1]
    S[table-format=1.4]
    S[table-format=1.4]
    S[table-format=1.4]}
\toprule
\textbf{Method}
  & {\textbf{PPL} $\downarrow$}
  & {\textbf{$\Delta$PPL\%}}
  & {\textbf{SR} $\downarrow$}
  & {\textbf{CHR} $\uparrow$}
  & {\textbf{Ent (bits)} $\uparrow$} \\
\midrule
Baseline
  & 245.297 & {---} & 0.7087 & 0.5366 & 1.9558 \\
\midrule
Soft ($\lambda=0.01$)  & 243.952 & -0.5 & 0.6702 & 0.5731 & 1.9459 \\
Soft ($\lambda=0.05$)  & 245.609 & +0.1 & 0.5573 & 0.6805 & 1.9522 \\
\rowcolor{bestcol}
Soft ($\lambda=0.10$)  & 247.582 & +0.9 & 0.4756 & 0.7450 & 1.9404 \\
Soft ($\lambda=0.20$)  & 249.999 & +1.9 & 0.3935 & 0.8156 & 1.9214 \\
Soft ($\lambda=0.50$)  & 254.771 & +3.9 & 0.2952 & 0.8692 & 1.9517 \\
\midrule
Hard ($W=2$)           & 258.996 & +5.6 & 0.4991 & 0.7576 & 1.9513 \\
Hard ($W=4$)           & 254.490 & +3.7 & 0.6094 & 0.6783 & 1.9459 \\
Hard ($W=8$)           & 248.116 & +1.1 & 0.6633 & 0.6036 & 1.9392 \\
\midrule
Soft-Hard ($\lambda=0.1$, $\alpha=0.05$, $W=4$)
                       & 247.942 & +1.1 & 0.4716 & 0.7561 & 1.9250 \\
Soft-Hard ($\lambda=0.1$, $\alpha=0.10$, $W=4$)
                       & 248.120 & +1.2 & 0.4478 & 0.7725 & 1.9327 \\
Soft-Hard ($\lambda=0.1$, $\alpha=0.50$, $W=4$)
                       & 251.264 & +2.4 & 0.3397 & 0.8426 & 1.9287 \\
Soft-Hard ($\lambda=0.1$, $\alpha=1.00$, $W=4$)
                       & 253.986 & +3.5 & 0.3342 & 0.8581 & 1.9451 \\
\midrule
ReMoE ($\lambda=0.1$)  & 245.034 & -0.1 & 0.7054 & 0.5397 & 1.9518 \\
Oracle-MoE             & 257.760 & +5.1 & 0.3907 & 0.8441 & 1.9378 \\
\bottomrule
\end{tabular}
\end{table}

\begin{table}[t]
\small
\centering
\caption{%
  Results on WikiText-2 \textbf{(Medium model, $\approx$22M parameters, 4 experts,
  top-2 routing)}.
  Columns as in \cref{tab:results_small}.
  \colorbox{bestcol}{Highlighted} row is the recommended operating point.
}
\label{tab:results_medium}
\setlength{\tabcolsep}{5pt}
\renewcommand{\arraystretch}{1.15}
\begin{tabular}{l
    S[table-format=3.3]
    S[table-format=+1.1]
    S[table-format=1.4]
    S[table-format=1.4]
    S[table-format=1.4]}
\toprule
\textbf{Method}
  & {\textbf{PPL} $\downarrow$}
  & {\textbf{$\Delta$PPL\%}}
  & {\textbf{SR} $\downarrow$}
  & {\textbf{CHR} $\uparrow$}
  & {\textbf{Ent (bits)} $\uparrow$} \\
\midrule
Baseline
  & 274.673 & {---} & 0.7127 & 0.5376 & 1.9694 \\
\midrule
Soft ($\lambda=0.01$)  & 270.230 & -1.6 & 0.6796 & 0.5734 & 1.9868 \\
Soft ($\lambda=0.05$)  & 263.468 & -4.1 & 0.5382 & 0.7039 & 1.9751 \\
Soft ($\lambda=0.10$)  & 269.104 & -2.0 & 0.4329 & 0.7795 & 1.9828 \\
Soft ($\lambda=0.20$)  & 268.789 & -2.1 & 0.3675 & 0.8323 & 1.9777 \\
\rowcolor{bestcol}
Soft ($\lambda=0.50$)  & 272.314 & -0.9 & 0.2906 & 0.8817 & 1.9625 \\
\midrule
Hard ($W=2$)           & 293.253 & +6.8 & 0.4724 & 0.7978 & 1.9654 \\
Hard ($W=4$)           & 293.059 & +6.7 & 0.6158 & 0.6860 & 1.9525 \\
Hard ($W=8$)           & 281.151 & +2.4 & 0.6825 & 0.5920 & 1.9561 \\
\midrule
Soft-Hard ($\lambda=0.1$, $\alpha=0.05$, $W=4$)
                       & 270.432 & -1.5 & 0.4124 & 0.7970 & 1.9736 \\
Soft-Hard ($\lambda=0.1$, $\alpha=0.10$, $W=4$)
                       & 270.898 & -1.4 & 0.3918 & 0.8156 & 1.9786 \\
Soft-Hard ($\lambda=0.1$, $\alpha=0.50$, $W=4$)
                       & 275.580 & +0.3 & 0.3245 & 0.8544 & 1.9721 \\
Soft-Hard ($\lambda=0.1$, $\alpha=1.00$, $W=4$)
                       & 275.861 & +0.4 & 0.2735 & 0.8892 & 1.9574 \\
\midrule
ReMoE ($\lambda=0.1$)  & 274.101 & -0.2 & 0.7093 & 0.5398 & 1.9684 \\
Oracle-MoE             & 278.567 & +1.4 & 0.3162 & 0.8862 & 1.9584 \\
\bottomrule
\end{tabular}
\end{table}

\paragraph{Soft consistency improves or preserves perplexity at low $\lambda$.}
On the small model, Soft $\lambda{=}0.01$ reduces perplexity relative to the
Baseline ($245.3 \to 243.9$, $-0.5\%$), and Soft $\lambda{=}0.05$ matches
the Baseline within $+0.1\%$. The effect is stronger on the medium model, where
Soft $\lambda{=}0.05$ yields the best perplexity of any method ($263.5$,
$-4.1\%$ relative to Baseline). This is a notable finding: at moderate $\lambda$,
the consistency loss acts as a beneficial regulariser rather than a quality
penalty, shaping representations that are both more locally coherent and more
predictive. Only at aggressive $\lambda \geq 0.2$ does a clear quality--locality
trade-off emerge.

\paragraph{Switch rate is substantially reduced across both scales.}
Soft $\lambda{=}0.5$ reduces SR from $0.71$ to $0.30$ on the small model and
from $0.71$ to $0.29$ on the medium model --- a consistent $\sim$59\% reduction
at both scales. The smooth monotone relationship between $\lambda$ and SR
(\cref{fig:pareto}) means practitioners can select an operating point to match
their hardware memory budget without retraining from scratch.

\paragraph{ReMoE fails to reduce switch rate at either scale.}
Post-hoc router fine-tuning achieves SR of $0.705$ on the small model and
$0.709$ on the medium model, compared to baselines of $0.709$ and $0.713$
respectively --- a negligible change of less than $0.5\%$ at both scales.
The per-layer analysis (\cref{tab:perlayer_small,tab:perlayer_medium}) makes
this even starker: ReMoE's per-layer switch rates are virtually identical to
the Baseline at every layer. This confirms the central claim of this paper ---
that routing locality cannot be meaningfully retrofitted post-hoc, because the
expert representations themselves were never trained to support it.

\paragraph{Hard baseline achieves locality at excessive quality cost.}
Hard ($W{=}2$) reduces SR to $0.499$ on the small model but at a $+5.6\%$
perplexity cost --- more than five times the cost of Soft $\lambda{=}0.1$ at
a comparable switch rate. More strikingly, larger windows do \emph{not} improve
locality: Hard ($W{=}8$) produces higher switch rates than Hard ($W{=}2$) at
every layer on both models, suggesting that the logit bias becomes ineffective
when the window is too wide to maintain consistent pressure. The Hard baseline
is therefore both expensive in quality and unreliable in locality, motivating
the soft differentiable approach.

\paragraph{Soft-Hard variant provides the best locality at moderate cost.}
On the medium model, Soft-Hard ($\lambda{=}0.1$, $\alpha{=}1.0$, $W{=}4$)
achieves the lowest switch rate of any method ($0.274$) with only $+0.4\%$
perplexity cost, and the highest cache hit rate ($0.889$). The per-layer
profile shows L1 dropping to $0.108$ --- an $85\%$ reduction from the Baseline
--- while perplexity remains within $1\%$ of the Baseline. This confirms that
the anchor constraint and the soft penalty are complementary: the soft loss
prevents local switching while the anchor prevents long-range drift within
a window.

\paragraph{Expert utilisation is preserved throughout.}
Utilisation entropy remains above $1.92$ bits across all \method{} variants
on both models, out of a maximum of $\log_2 4 = 2.0$ bits. Even at $\lambda{=}0.5$,
where SR is reduced by nearly $60\%$, all four experts remain actively used.
The consistency loss does not cause the model to collapse routing to a single
expert --- it redistributes \emph{when} experts are used, not \emph{which}
experts exist.

\subsection{Pareto Frontier and Inference-Time Impact}

\method{} requires no modification to the forward pass at inference time ---
the trained router naturally produces more temporally consistent gate
distributions, and any standard caching strategy (LRU, LFU, or learned)
benefits from the improved locality without additional engineering effort.

\begin{figure}[htbp]
  \centering
  \begin{subfigure}[b]{0.8\textwidth}
    \centering
    \includegraphics[width=\textwidth]{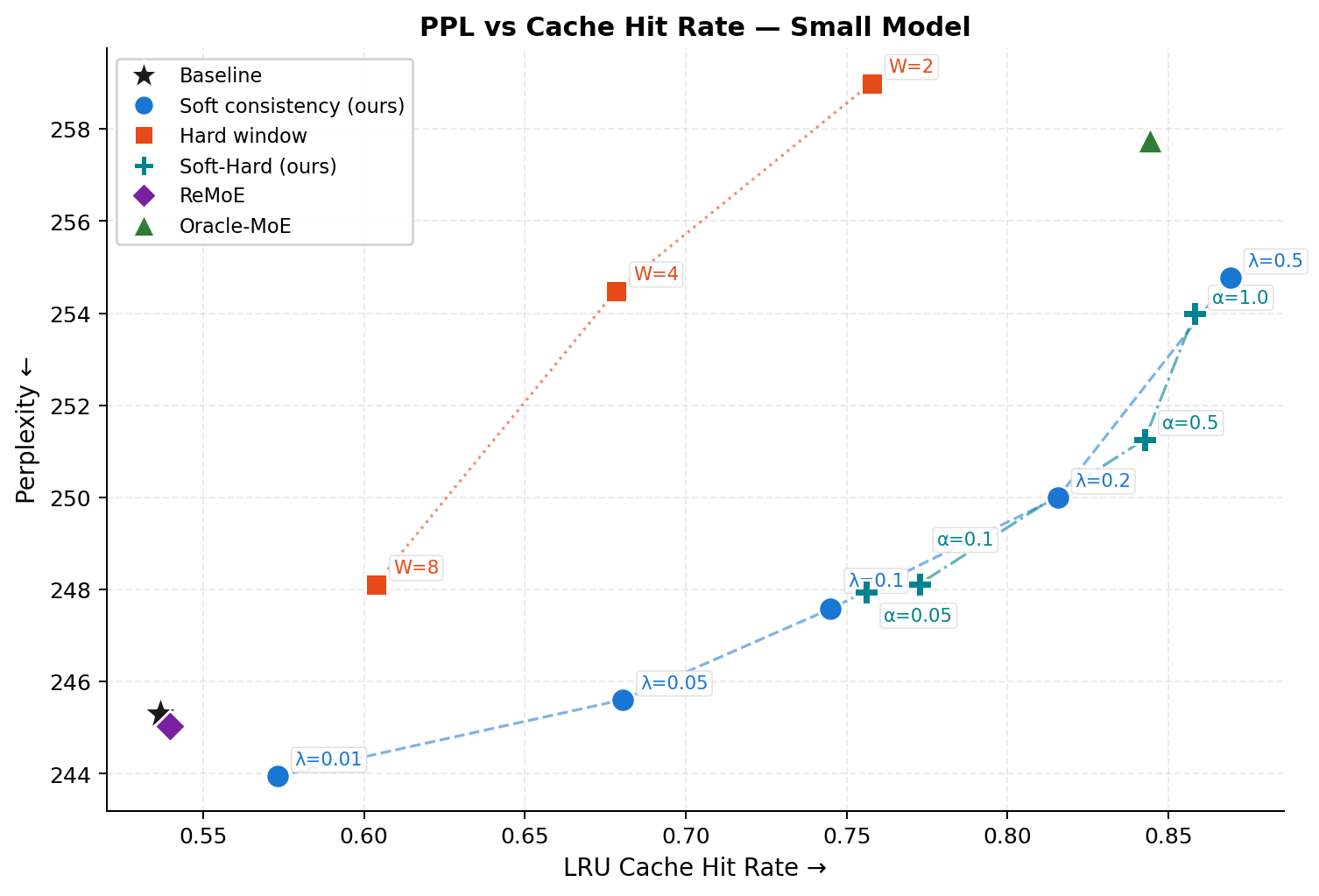}
    \caption{Small model (${\sim}$8.8M params)}
  \end{subfigure}
  \hfill
  \begin{subfigure}[b]{0.8\textwidth}
    \centering
    \includegraphics[width=\textwidth]{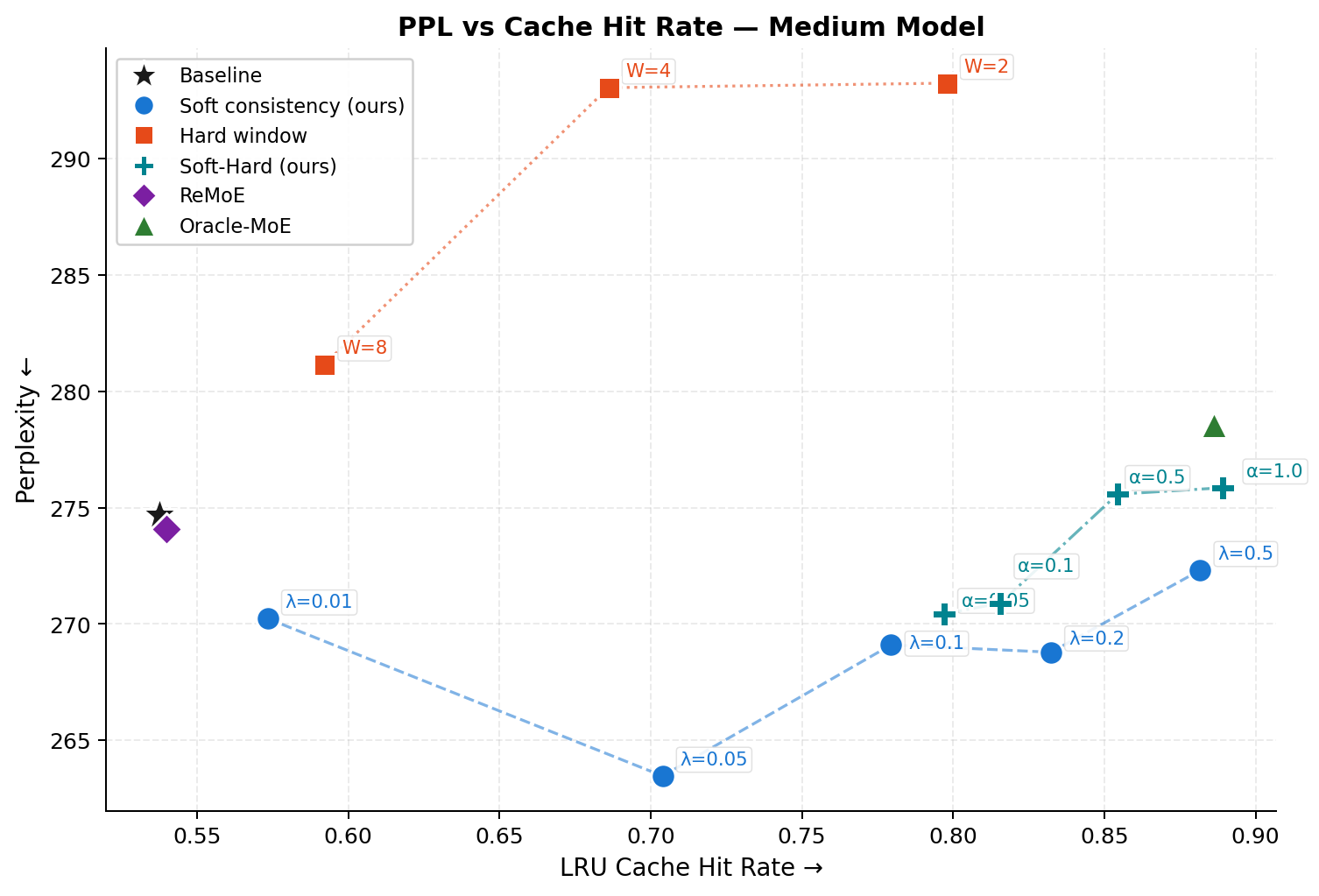}
    \caption{Medium model (${\sim}$22M params)}
  \end{subfigure}
  \caption{Quality--locality Pareto frontier (cache hit rate vs.\ perplexity) on
  WikiText-2 for small and medium models. Each point represents one method
  configuration; higher cache hit rate and lower perplexity are both desirable
  (upper-left is better). \method{} Soft and Soft-Hard variants trace a smooth
  Pareto frontier, consistently achieving higher cache hit rates at comparable or
  better perplexity than the Hard baseline, ReMoE, and Oracle-MoE. ReMoE
  fails to meaningfully improve cache hit rate at either scale, clustering near
  the Baseline point.}
  \label{fig:pareto}
\end{figure}

Figure~\ref{fig:pareto} illustrates the quality--locality Pareto frontier.
\method{} Soft and Soft-Hard variants strictly dominate ReMoE across all
operating points: because ReMoE fails to move SR meaningfully, it offers
no cache efficiency benefit whatsoever despite the additional fine-tuning
cost. Oracle-MoE achieves competitive locality but requires architectural
changes to the router input; \method{} matches or exceeds its cache hit
rate without any structural modification. The Hard baseline traces a separate,
dominated frontier --- high locality is achievable but only at perplexity
costs that make the model uncompetitive.

The cache impact is directly quantified by our LRU simulations ($C{=}2$
expert slots). On the small model, \method{} at $\lambda{=}0.1$ reduces SR
from $0.71$ to $0.47$, raising CHR from $0.54$ to $0.75$ --- a $1.82\times$
reduction in cache misses. The effect is more pronounced on the medium model:
at $\lambda{=}0.5$, SR drops from $0.71$ to $0.29$, raising CHR from $0.54$
to $0.88$ --- a $3.92\times$ reduction in cache misses, translating directly
into lower per-token latency on bandwidth-limited edge hardware.

\subsection{Ablation Studies}

\begin{table}[t]
\small
\centering
\caption{%
  Per-layer expert switch rate on WikiText-2
  \textbf{(Medium model, $\approx$22M parameters)}.
  SR per layer (lower is better). Layer 0 is closest to the input.
}
\label{tab:perlayer_medium}
\setlength{\tabcolsep}{10pt}
\renewcommand{\arraystretch}{1.15}
\begin{tabular}{l
    S[table-format=1.3]
    S[table-format=1.3]
    S[table-format=1.3]
    S[table-format=1.3]}
\toprule
\textbf{Method}
  & {\textbf{L0} $\downarrow$}
  & {\textbf{L1} $\downarrow$}
  & {\textbf{L2} $\downarrow$}
  & {\textbf{L3} $\downarrow$} \\
\midrule
Baseline                                         & 0.732 & 0.739 & 0.733 & 0.646 \\
\midrule
Soft ($\lambda=0.01$)                            & 0.722 & 0.717 & 0.660 & 0.619 \\
Soft ($\lambda=0.05$)                            & 0.674 & 0.577 & 0.557 & 0.346 \\
Soft ($\lambda=0.10$)                            & 0.642 & 0.423 & 0.367 & 0.299 \\
Soft ($\lambda=0.20$)                            & 0.617 & 0.317 & 0.291 & 0.245 \\
Soft ($\lambda=0.50$)                            & 0.549 & 0.177 & 0.232 & 0.205 \\
\midrule
Hard ($W=2$)                                     & 0.519 & 0.462 & 0.417 & 0.492 \\
Hard ($W=4$)                                     & 0.621 & 0.683 & 0.602 & 0.557 \\
Hard ($W=8$)                                     & 0.707 & 0.716 & 0.714 & 0.593 \\
\midrule
Soft-Hard ($\lambda=0.1$, $\alpha=0.05$, $W=4$) & 0.638 & 0.390 & 0.378 & 0.245 \\
Soft-Hard ($\lambda=0.1$, $\alpha=0.10$, $W=4$) & 0.637 & 0.326 & 0.324 & 0.279 \\
Soft-Hard ($\lambda=0.1$, $\alpha=0.50$, $W=4$) & 0.620 & 0.181 & 0.299 & 0.197 \\
Soft-Hard ($\lambda=0.1$, $\alpha=1.00$, $W=4$) & 0.576 & 0.108 & 0.215 & 0.196 \\
\midrule
ReMoE ($\lambda=0.1$)                            & 0.731 & 0.737 & 0.733 & 0.636 \\
Oracle-MoE                                       & 0.513 & 0.166 & 0.306 & 0.280 \\
\bottomrule
\end{tabular}
\end{table}

\paragraph{Per-layer consistency profiles.}
Tables~\ref{tab:perlayer_small} and~\ref{tab:perlayer_medium} show per-layer
switch rates across all methods. Two consistent patterns emerge across both
model sizes. First, L0 is the most resistant layer: even at $\lambda{=}0.5$,
L0 SR remains above $0.49$ on both models, compared to reductions to $0.22$
and below at L1--L3. This is expected --- L0 processes token embeddings before
any contextual mixing via attention, making adjacent tokens intrinsically more
dissimilar at that layer. Second, the consistency loss distributes its effect
non-uniformly: L1 and L2 see the largest absolute reductions, suggesting these
layers are most amenable to locality regularisation once some contextual
integration has occurred. The Soft-Hard variant at $\alpha{=}1.0$ achieves
the single lowest per-layer SR in either table: L1 $= 0.108$ on the medium
model, an $85\%$ reduction from the Baseline value of $0.739$.

\paragraph{$\lambda$ sensitivity.}
The medium model is substantially more robust to large $\lambda$ than the
small model. On the small model, $\lambda{=}0.5$ incurs $+3.9\%$ PPL; on
the medium model the same $\lambda$ incurs only $-0.9\%$ (an improvement).
This scaling behaviour is encouraging: as model capacity grows, the
consistency constraint becomes easier to satisfy without sacrificing
representational quality, suggesting \method{} will become more effective,
not less, at larger scales.

\section{Discussion}
\label{sec:discussion}

\paragraph{When does \method{} help most?}
The consistency loss is most beneficial when two conditions hold simultaneously:
the target hardware has severely constrained fast memory relative to expert size,
making each cache miss expensive, and the target domain has natural local coherence
--- long-form text, code, conversations, documents --- that the consistency loss
can reinforce. Our results suggest a third condition that was not anticipated:
\method{} is more effective at larger model scales. On the medium model, moderate
$\lambda$ values actually \emph{improve} perplexity relative to the baseline,
suggesting that larger expert networks have sufficient capacity to satisfy the
routing constraint without sacrificing representational quality. This is an
encouraging sign for deployment at scale.

\paragraph{Why post-hoc fine-tuning fails.}
The most striking empirical finding is the complete ineffectiveness of ReMoE-style
post-hoc router fine-tuning. Across both model sizes and all four layers, ReMoE
produces switch rates within $0.5\%$ of the Baseline --- a negligible change
despite additional compute. This is not a failure of the fine-tuning procedure
itself but a fundamental consequence of the representation mismatch problem: the
expert weight matrices were trained without any locality incentive, and the
hidden representations they produce do not cluster in a way that supports
temporally stable routing. Fine-tuning only the router cannot overcome this
because the router's input space is fixed. \method{} avoids this by shaping both
representations and routing decisions jointly from the first training step.

\paragraph{The L0 barrier.}
Across all methods and both model sizes, layer L0 is consistently the most
resistant to locality improvement. Even at $\lambda{=}0.5$, L0 switch rate
remains above $0.49$, while L1--L3 drop to $0.17$--$0.24$. We attribute this
to the nature of L0 inputs: token embeddings carry token-specific identity
information and have not yet been mixed with contextual information via
attention. Adjacent tokens are therefore genuinely dissimilar at L0 in a
way that the consistency loss cannot easily overcome without large perplexity
cost. This suggests that future work targeting L0 specifically --- perhaps
through embedding-level smoothing or subword-aware routing --- could yield
further locality gains.

\paragraph{Relationship to language structure.}
The non-uniform distribution of locality gains across layers is consistent
with the view that higher layers represent more abstract, semantically stable
information while lower layers process surface-level token features. \method{}
effectively asks lower layers to behave more like higher layers from a routing
perspective. The fact that this is achievable at moderate $\lambda$ without
quality loss suggests that the routing surface in these layers is more flexible
than the representations themselves --- the router can be steered toward
consistency without the underlying features needing to change dramatically.
Consistent with this, L1 SR drops from $0.739$ to $0.177$ at $\lambda{=}0.5$
on the medium model while PPL simultaneously \emph{improves} by $0.9\%$ ---
evidence that the routing surface at L1 is genuinely more flexible than the
representations it operates on.

\paragraph{Towards cross-layer routing consistency.}
The consistency loss proposed in this paper operates within each layer across
consecutive tokens, which is most directly beneficial during prefill and batched
decode. For single-token autoregressive decoding --- the dominant inference mode
on memory-constrained edge devices --- the forward pass traverses all layers
sequentially for each token, making cross-layer memory access the primary
bottleneck. A complementary cross-layer consistency loss that encourages the
same expert to be selected at consecutive layers for the same token would
directly target this bottleneck. However, such a loss only reduces cache misses
if expert weights are shared across layers, since routing consistently to expert
$i$ at layers $\ell$ and $\ell+1$ only avoids a memory load when
$\calE_i^{(\ell)} = \calE_i^{(\ell+1)}$. In standard MoE models with independent
per-layer expert matrices, cross-layer routing consistency provides no cache
benefit. We provide a full analysis of this access pattern, the parameter sharing
requirement, and a proposed combined training objective in \cref{app:crosslayer}.

\paragraph{Limitations.}
The consistency loss penalises switches uniformly regardless of whether a switch
occurs at a natural semantic boundary --- a sentence break, topic shift, or
dialogue turn --- or in the middle of a coherent span. At a sentence boundary,
switching experts is semantically appropriate and penalising it unnecessarily
constrains the model. A boundary-aware variant that gates the consistency loss
using punctuation or BOS signals would relax the penalty precisely where
switching is warranted and strengthen it where it is not. We expect this would
improve the quality--locality trade-off particularly at higher $\lambda$ values
where the current formulation begins to hurt perplexity. We leave this to
future work.

\section{Conclusion}
\label{sec:conclusion}

We introduced \method{}, a training-time routing consistency loss for
Mixture-of-Experts language models that directly optimises for temporal locality
in expert activation patterns. By penalising abrupt expert switches between
adjacent tokens during pretraining, \method{} produces routers that are
intrinsically locality-aware without requiring any architectural modifications
or post-hoc adaptation. Experiments on small and medium MoE models demonstrate
that \method{} reduces the expert switch rate by up to \textbf{59\%} while
simultaneously \emph{improving} perplexity on the medium model, and reduces
cache misses by up to $\mathbf{3.92\times}$ --- a result that stands in stark
contrast to post-hoc router fine-tuning, which fails to meaningfully reduce
switch rate at either scale, confirming that routing locality cannot be
retrofitted and must be instilled during training.

These results instantiate a broader principle: inference-time memory access
patterns are a legitimate training objective. Just as quantisation-aware
training~\citep{jacob2018quantization} has become standard practice by
demonstrating that training with deployment constraints in the loop consistently
outperforms post-hoc compression, we argue that locality-aware training should
be the default starting point for MoE models intended for memory-constrained
deployment. The post-hoc stack --- caching heuristics, router fine-tuning,
offloading schedulers --- remains useful as a complement, but not as a
substitute for a model that was never asked to be cache-friendly in the
first place.

\paragraph{Future work.}
Promising directions include: scaling \method{} to larger MoE models
(1B+ parameters) where the regularisation benefit observed at medium scale
may be even more pronounced; boundary-aware consistency losses that relax
the penalty at sentence and paragraph boundaries; combining \method{} with
structured expert grouping so that switches within a group are cheaper than
switches across groups; and investigating whether experts trained with \method{}
develop more interpretable and topically coherent specialisations, given that
they process longer contiguous spans of semantically related tokens --- a
testable prediction that consistent routing implies more coherent and persistent
expert assignments across a document.

\bibliographystyle{plainnat}
\bibliography{refs}

\appendix

\section{Reproducibility}

All code, configuration files, and dependency versions are committed to the project
repository. Exact package versions are specified in \texttt{uv.lock}. The WikiText-2
dataset is downloaded from the HuggingFace Hub
(\texttt{Salesforce/wikitext}, split \texttt{wikitext-2-raw-v1}) and cached locally as
NumPy arrays. The full experiment pipeline is reproducible by running:
\begin{verbatim}
    uv sync
    uv run python download_data.py
    bash run_experiments.sh
\end{verbatim}

\section{Switch Rate Pareto Frontier}

\begin{figure}[htbp]
  \centering
  \begin{subfigure}[b]{0.8\textwidth}
    \centering
    \includegraphics[width=\textwidth]{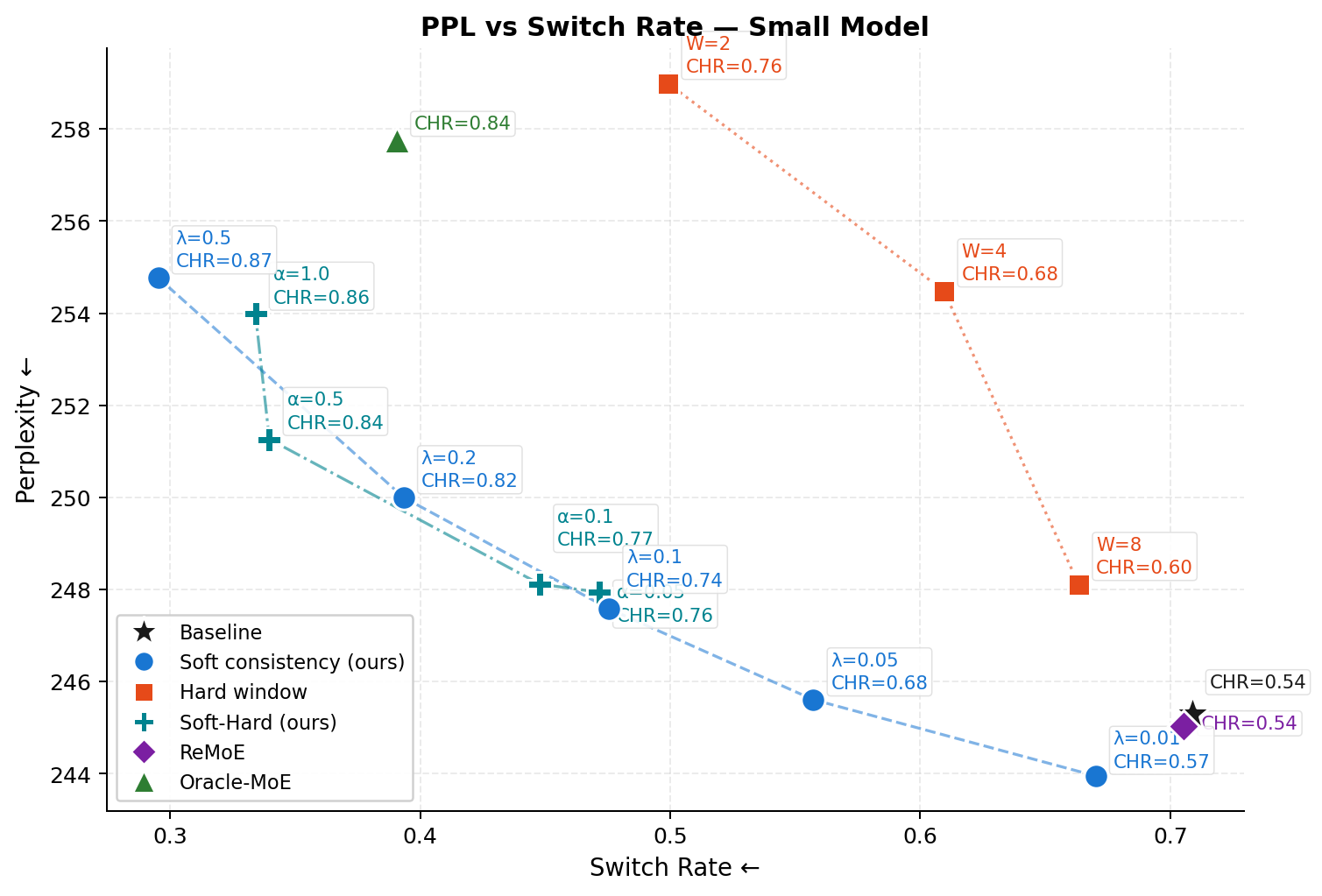}
    \caption{Small model (${\sim}$8.8M params)}
  \end{subfigure}
  \hfill
  \begin{subfigure}[b]{0.8\textwidth}
    \centering
    \includegraphics[width=\textwidth]{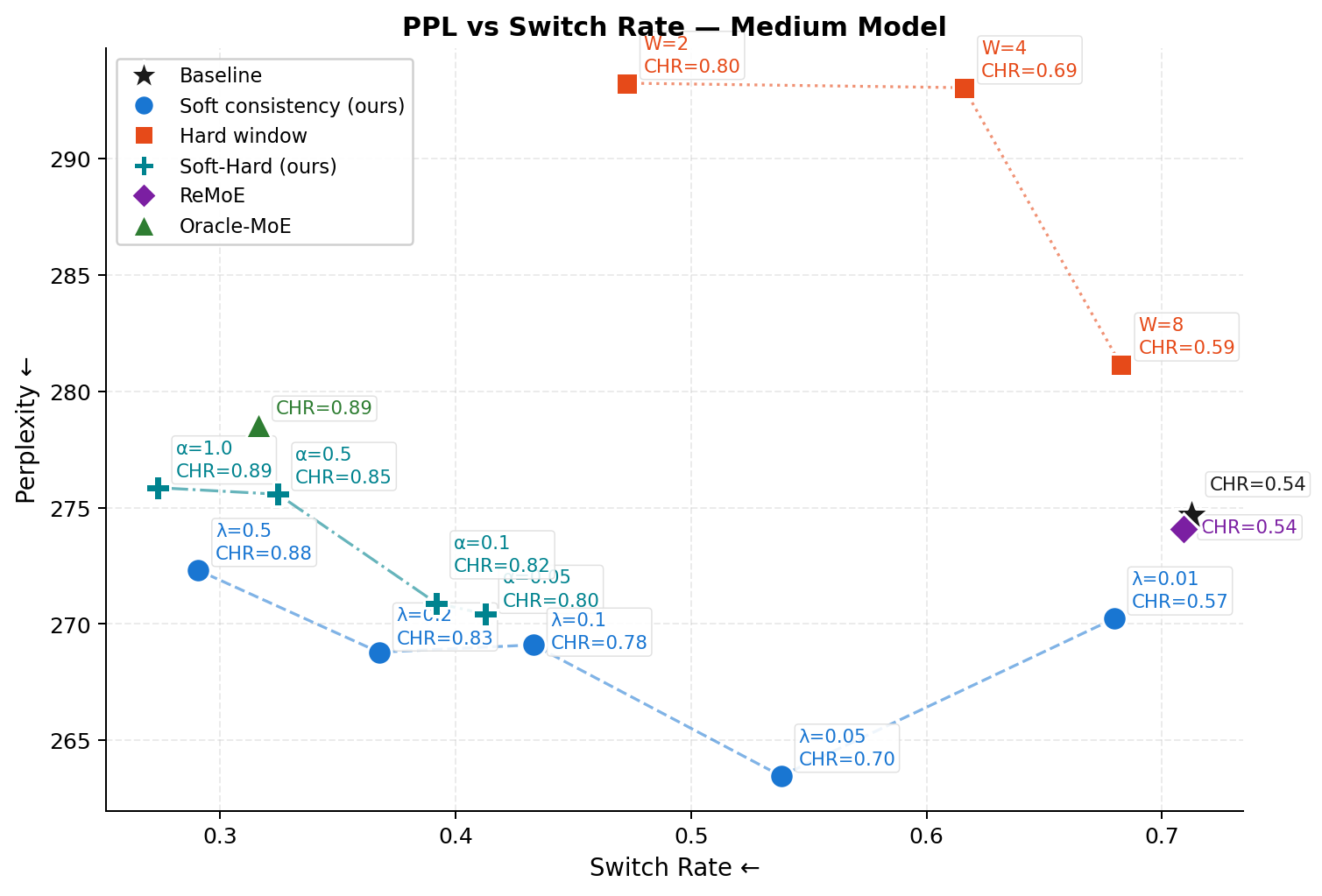}
    \caption{Medium model (${\sim}$22M params)}
  \end{subfigure}
  \caption{Quality--locality Pareto frontier (switch rate vs.\ perplexity) on
  WikiText-2 for small and medium models. Each point represents one method
  configuration; lower switch rate and lower perplexity are both desirable
  (lower-left is better). \method{} Soft and Soft-Hard variants trace a smooth
  Pareto frontier, consistently achieving lower switch rates at comparable or
  better perplexity than the Hard baseline, ReMoE, and Oracle-MoE. ReMoE
  fails to meaningfully reduce switch rate at either scale, clustering near
  the Baseline point.}
  \label{fig:pareto_sr}
\end{figure}

Figure~\ref{fig:pareto_sr} shows the quality--locality Pareto frontier using switch
rate as the locality metric, complementing the cache hit rate frontier in the main
text.

\section{Per-Layer Switch Rates: Small Model}

\begin{table}[h]
\small
\centering
\caption{%
  Per-layer expert switch rate on WikiText-2
  \textbf{(Small model, $\approx$8.8M parameters)}.
  SR per layer (lower is better). Layer 0 is closest to the input.
}
\label{tab:perlayer_small}
\setlength{\tabcolsep}{10pt}
\renewcommand{\arraystretch}{1.15}
\begin{tabular}{l
    S[table-format=1.3]
    S[table-format=1.3]
    S[table-format=1.3]
    S[table-format=1.3]}
\toprule
\textbf{Method}
  & {\textbf{L0} $\downarrow$}
  & {\textbf{L1} $\downarrow$}
  & {\textbf{L2} $\downarrow$}
  & {\textbf{L3} $\downarrow$} \\
\midrule
Baseline                                        & 0.716 & 0.714 & 0.715 & 0.691 \\
\midrule
Soft ($\lambda=0.01$)                           & 0.680 & 0.684 & 0.681 & 0.636 \\
Soft ($\lambda=0.05$)                           & 0.653 & 0.597 & 0.459 & 0.520 \\
Soft ($\lambda=0.10$)                           & 0.641 & 0.546 & 0.326 & 0.389 \\
Soft ($\lambda=0.20$)                           & 0.571 & 0.428 & 0.304 & 0.271 \\
Soft ($\lambda=0.50$)                           & 0.488 & 0.234 & 0.220 & 0.239 \\
\midrule
Hard ($W=2$)                                    & 0.563 & 0.538 & 0.531 & 0.363 \\
Hard ($W=4$)                                    & 0.628 & 0.679 & 0.600 & 0.531 \\
Hard ($W=8$)                                    & 0.670 & 0.729 & 0.676 & 0.578 \\
\midrule
Soft-Hard ($\lambda=0.1$, $\alpha=0.05$, $W=4$) & 0.647 & 0.532 & 0.327 & 0.380 \\
Soft-Hard ($\lambda=0.1$, $\alpha=0.10$, $W=4$) & 0.622 & 0.521 & 0.325 & 0.324 \\
Soft-Hard ($\lambda=0.1$, $\alpha=0.50$, $W=4$) & 0.600 & 0.281 & 0.270 & 0.208 \\
Soft-Hard ($\lambda=0.1$, $\alpha=1.00$, $W=4$) & 0.585 & 0.264 & 0.235 & 0.252 \\
\midrule
ReMoE ($\lambda=0.1$)                           & 0.715 & 0.712 & 0.711 & 0.684 \\
Oracle-MoE                                      & 0.676 & 0.229 & 0.290 & 0.367 \\
\bottomrule
\end{tabular}
\end{table}

\section{Cross-Layer Routing Consistency: A Forward-Looking Analysis}
\label{app:crosslayer}

\subsection{The Decode-Time Memory Access Pattern}

The consistency loss proposed in the main paper operates \emph{within} a layer
across consecutive tokens:
\begin{equation}
  \calL_{\text{cons}}^{(\ell)} = \frac{1}{T-1}\sum_{t=2}^{T}
  \bigl\| \mathbf{g}_t^{(\ell)} - \mathbf{g}_{t-1}^{(\ell)} \bigr\|_2^2.
\end{equation}
This formulation is well-motivated for \emph{prefill} (where many tokens are
processed in parallel, one layer at a time) and for batched decode. For
single-token autoregressive decoding --- the dominant inference mode on
memory-constrained edge devices --- the memory access pattern is different.
The forward pass processes one token at a time and traverses all layers
sequentially:
\begin{equation}
  \text{token } t: \quad
  \calE^{(0)} \;\to\; \calE^{(1)} \;\to\; \calE^{(2)} \;\to\; \cdots \;\to\; \calE^{(L-1)},
\end{equation}
so consecutive memory accesses are \emph{across layers for the same token},
not across tokens for the same layer. The access sequence for two consecutive
tokens under single-token decode is:

\begin{center}
\begin{tabular}{lll}
\toprule
\textbf{Step} & \textbf{Token} & \textbf{Expert loaded} \\
\midrule
1 & $t$   & $\calE^{(0)}_{\sigma_t^{(0)}}$ \\
2 & $t$   & $\calE^{(1)}_{\sigma_t^{(1)}}$ \\
3 & $t$   & $\calE^{(2)}_{\sigma_t^{(2)}}$ \\
4 & $t$   & $\calE^{(3)}_{\sigma_t^{(3)}}$ \\
5 & $t+1$ & $\calE^{(0)}_{\sigma_{t+1}^{(0)}}$ \\
6 & $t+1$ & $\calE^{(1)}_{\sigma_{t+1}^{(1)}}$ \\
$\vdots$ & $\vdots$ & $\vdots$ \\
\bottomrule
\end{tabular}
\end{center}

where $\sigma_t^{(\ell)}$ denotes the expert index selected for token $t$ at
layer $\ell$. Each step requires loading a potentially different expert weight
matrix. Under this access pattern, the token-level consistency loss reduces
misses at step 5 relative to step 1 (same layer, adjacent tokens) but does
nothing to reduce misses at step 2 relative to step 1 (adjacent layers, same
token).

\subsection{Cross-Layer Consistency and the Parameter Sharing Requirement}

A natural extension is a \emph{cross-layer consistency loss} that encourages
the same expert to be selected at consecutive layers for the same token:
\begin{equation}
  \calL_{\text{cross}} = \frac{1}{T}\sum_{t=1}^{T}\sum_{\ell=1}^{L-1}
  \bigl\| \mathbf{g}_t^{(\ell)} - \mathbf{g}_t^{(\ell+1)} \bigr\|_2^2.
  \label{eq:cross_loss}
\end{equation}
However, there is a critical prerequisite: cross-layer routing consistency only
reduces cache misses if the \emph{same expert index at different layers refers
to the same weight tensor in memory}. In a standard MoE model where each layer
maintains its own independent set of expert matrices
$\{\calE_i^{(\ell)}\}_{i=1}^{N}$, expert 2 at layer $\ell$ and expert 2 at
layer $\ell+1$ are entirely different weight tensors. Routing consistently to
expert 2 across both layers still requires two separate memory loads --- no
cache benefit is obtained.

Cross-layer consistency therefore becomes practically useful only under one of
the following architectural conditions:

\begin{enumerate}
  \item \textbf{Full cross-layer weight sharing.} Expert $i$ is the same weight
        matrix at every layer: $\calE_i^{(\ell)} = \calE_i$ for all $\ell$.
        Consistent routing to expert $i$ across all layers requires only a
        single load, amortised over $L$ layers of computation.

  \item \textbf{Grouped weight sharing.} Layers are partitioned into blocks of
        size $K$ (e.g. $K=2$ or $K=4$), and all layers within a block share the
        same expert pool. Consistent routing within a block reduces loads;
        routing across block boundaries still requires new loads.

  \item \textbf{Universal expert pool.} All layers draw from one shared set of
        expert matrices --- a design explored in recent parameter-efficient
        architectures. Any cross-layer routing consistency directly translates
        to cache reuse.
\end{enumerate}

\subsection{The Combined Objective}

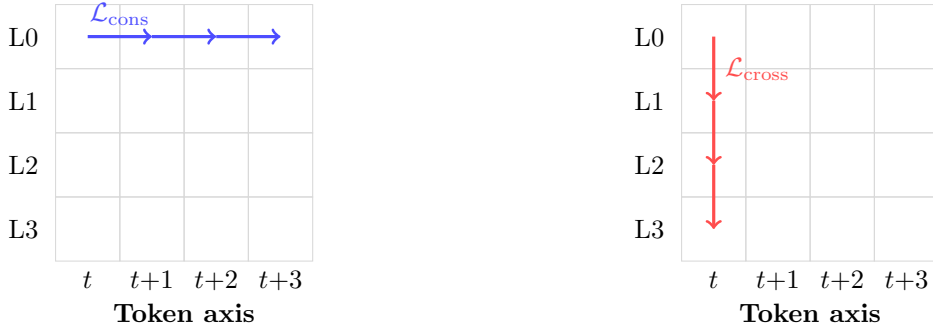
\begin{figure}[h]
  \centering
  \begin{subfigure}[b]{0.48\textwidth}
    \centering
    \begin{tikzpicture}[scale=0.85]
      \foreach \t in {0,1,2,3} {
        \foreach \l in {0,1,2,3} {
          \draw[gray!30] (\t,\l) rectangle (\t+1,\l+1);
        }
      }
      \draw[blue!70, very thick, ->] (0.5,3.5) -- (1.5,3.5)
        node[midway, above, font=\small] {$\calL_{\text{cons}}$};
      \draw[blue!70, very thick, ->] (1.5,3.5) -- (2.5,3.5);
      \draw[blue!70, very thick, ->] (2.5,3.5) -- (3.5,3.5);
      \foreach \t/\name in {0/$t$,1/$t{+}1$,2/$t{+}2$,3/$t{+}3$} {
        \node[font=\small] at (\t+0.5, -0.3) {\name};
      }
      \foreach \l/\name in {0/L3,1/L2,2/L1,3/L0} {
        \node[font=\small] at (-0.5, \l+0.5) {\name};
      }
      \node[font=\small\bfseries] at (2, -0.8) {Token axis};
    \end{tikzpicture}
    \caption{Token-level consistency (prefill / batched decode). Arrows show
    penalised pairs under $\calL_{\text{cons}}$ at layer L0.}
  \end{subfigure}
  \hfill
  \begin{subfigure}[b]{0.48\textwidth}
    \centering
    \begin{tikzpicture}[scale=0.85]
      \foreach \t in {0,1,2,3} {
        \foreach \l in {0,1,2,3} {
          \draw[gray!30] (\t,\l) rectangle (\t+1,\l+1);
        }
      }
      \draw[red!70, very thick, ->] (0.5,3.5) -- (0.5,2.5)
        node[midway, right, font=\small] {$\calL_{\text{cross}}$};
      \draw[red!70, very thick, ->] (0.5,2.5) -- (0.5,1.5);
      \draw[red!70, very thick, ->] (0.5,1.5) -- (0.5,0.5);
      \foreach \t/\name in {0/$t$,1/$t{+}1$,2/$t{+}2$,3/$t{+}3$} {
        \node[font=\small] at (\t+0.5, -0.3) {\name};
      }
      \foreach \l/\name in {0/L3,1/L2,2/L1,3/L0} {
        \node[font=\small] at (-0.5, \l+0.5) {\name};
      }
      \node[font=\small\bfseries] at (2, -0.8) {Token axis};
    \end{tikzpicture}
    \caption{Cross-layer consistency (single-token decode). Arrows show
    penalised pairs under $\calL_{\text{cross}}$ for token $t$.}
  \end{subfigure}
  \caption{The two memory access patterns and corresponding consistency
  objectives. Blue arrows (left) show token-level consistency targeting
  prefill and batched decode. Red arrows (right) show cross-layer consistency
  targeting single-token autoregressive decode. Both objectives are
  complementary and can be combined in \cref{eq:full_loss}.}
  \label{fig:access_patterns}
\end{figure}

Under a grouped weight sharing architecture with block size $K$, the full
memory-aware training objective combining token-level and cross-layer
consistency would be:
\begin{equation}
  \calL = \calL_{\text{CE}}
        + \lambda\,\calL_{\text{cons}}
        + \mu\,\calL_{\text{bal}}
        + \gamma\,\calL_{\text{cross}},
  \label{eq:full_loss}
\end{equation}
where $\gamma \geq 0$ controls the cross-layer consistency weight. The two
consistency terms are complementary and target different bottlenecks:
$\calL_{\text{cons}}$ reduces misses across tokens at the same layer (prefill
and batched decode), while $\calL_{\text{cross}}$ reduces misses across layers
for the same token (single-token decode). \Cref{fig:access_patterns} illustrates
the two access patterns and the corresponding consistency objectives.

\subsection{Limiting Case: Pre-defined Expert Trajectories}

The consistency objectives proposed in this paper nudge the router toward
stable routing decisions but still allow it to adapt freely. A natural
limiting case is to remove the router entirely and fix expert assignments
before training begins --- defining a deterministic \emph{trajectory}
$\sigma_t = (\sigma_t^{(0)}, \sigma_t^{(1)}, \ldots, \sigma_t^{(L-1)})$
through the network for every token $t$, where $\sigma_t^{(\ell)} \in
\{1,\ldots,N\}$ is the expert index at layer $\ell$.

The idea of fixed routing is not new: \citet{roller2021hash} show that
assigning tokens to experts via a deterministic hash function, with no
learned router, produces surprisingly competitive models. Our trajectory
framing differs in two key respects. First, assignments are coordinated
\emph{across layers} to define a complete network path per token, rather
than independent per-layer hashes. Second, trajectories are chosen to
maximise cache locality --- consecutive tokens are assigned the same
trajectory, so the entire sequence of expert weight loads for token $t$
is identical to that of token $t{-}1$, yielding zero cache misses within
a coherent span.

Under this design the cache miss rate during single-token autoregressive
decode is determined entirely by how often the trajectory changes between
consecutive tokens --- a quantity that is fully under the designer's
control rather than an emergent property of learned routing. In the extreme
case where all tokens in a document share the same trajectory, cache misses
are eliminated entirely at the cost of routing flexibility. This represents
the upper bound on locality achievable by any routing strategy, and serves
as a useful reference point for evaluating how close learned consistency
objectives like \method{} come to the theoretical optimum.

Whether a model trained with fixed locality-maximising trajectories can
match the quality of a model with learned routing is an open empirical
question. The hash routing results of \citet{roller2021hash} suggest it
is plausible, at least at small to moderate scale. We leave a systematic
investigation of this trade-off to future work.

\subsection{Implications and Future Work}

The analysis above identifies a gap in the current \method{} formulation: it
is most directly beneficial for prefill and batched decode, while single-token
autoregressive decode --- the dominant mode on memory-constrained edge devices
--- is better served by cross-layer consistency under a parameter-sharing
architecture. We leave the empirical validation of $\calL_{\text{cross}}$ and
its interaction with grouped weight sharing to future work, noting that the
training-time intervention principle extends naturally to this setting: just as
token-level locality must be instilled during training rather than retrofitted,
cross-layer locality requires architectural choices (parameter sharing) and
training objectives ($\calL_{\text{cross}}$) that are made from the outset.

\end{document}